# Methodological Rigour in Algorithm Application: An Illustration of Topic Modelling Algorithm

Malmi Amadoru
*University of Sydney*, malmi.amadoru@sydney.edu.au

# Methodological Rigour in Algorithm Application: An Illustration of Topic Modelling Algorithm

## Full research paper

**Malmi Amadoru**
Business Information Systems
University of Sydney Business School
Sydney, Australia
Email: malmi.amadoru@sydney.edu.au

## Abstract

The rise of advanced computational algorithms has opened new avenues for computationally intensive research approaches to theory development. However, the opacity of these algorithms and lack of transparency and rigour in their application pose methodological challenges, potentially undermining trust in research. The discourse on methodological rigour in this new genre of research is still emerging. Against this backdrop, I attempt to offer guidance on methodological rigour, particularly in the context of topic modelling algorithms. By illustrating the application of the structural topic modelling algorithm and presenting a set of guidelines, I discuss how to ensure rigour in topic modelling studies. Although the guidelines are for the application of topic modelling algorithms, they can be applied to other algorithms with context-specific adjustments. The guidelines are helpful, especially for novice researchers applying topic modelling, and editors and reviewers handling topic modelling manuscripts. I contribute to the literature on topic modelling and join the emerging dialogue on methodological rigour in computationally intensive theory construction research.

# 1 Introduction

The rapid development of advanced computational algorithms has expanded how research can be conducted. In recent years, scholars have acknowledged the potential of computationally intensive research approaches to theory development (Berente et al. 2019; Hannigan et al. 2019; Lindberg 2020; Miranda et al. 2022a; Shrestha et al. 2021). Computationally intensive theory construction (CITC) is an emerging genre of research that leverages algorithms often combined with other manual approaches such as qualitative methods to develop theory (Miranda et al. 2022a). Computationally derived patterns are core to this type of research as they form the basis for the theorising process. Scholars are using a variety of algorithms including topic modelling (Miranda et al. 2022b), sequence analysis (Lindberg et al. 2022), and agent-based modelling (Sturm et al. 2021). While there is huge potential for this emerging research genre, using algorithms in the research process poses new methodological challenges.

Algorithms are opaque by nature thus making it difficult to understand how an algorithm derives a specific output (Lebovitz et al. 2022). Often, researchers possess a limited understanding of an algorithm's inner workings, and this obscurity can make errors go unnoticed. In addition, algorithmic output is influenced by the data quality and hyperparameters specified by the researcher (Shrestha et al. 2021; Yang and Shami 2020). These factors make it difficult to understand how algorithms generate particular patterns and whether those patterns are plausible enough to derive knowledge claims. Yet, we are observing an influx of algorithms in research settings with limited transparency in computational procedures and a lack of methodological rigour (Günther and Joshi 2020). Such practices can potentially undermine trust in research because "*it is unlikely the quality of the whole will be high if the quality of the parts is not high*"(Weber 2012, p. 6). While established methodological genres such as qualitative, and mixed methods research have clear guidelines (e.g., Klein and Myers 1999; Venkatesh et al. 2013), directions for using algorithms in theory development are still emerging (e.g., simulation research guidelines by Dong 2022). Against this backdrop, I attempt to provide a set of guidelines for establishing methodological rigour for scholars who wish to use algorithms in research settings.

As algorithms are emerging rapidly, each demanding unique practices for rigour, I specifically narrow my focus to the topic modelling algorithms in this paper. Topic modelling algorithms are used to extract latent topics from a collection of documents (Blei 2012). They are being used widely by information systems and management scholars for theory-development purposes (Günther and Joshi 2020; Hannigan et al. 2019). For example, topic modelling has been used to understand IT innovation discourse (Miranda et al. 2022b) and social movements such as the Women's March (Syed and Silva 2023). Moreover, topic modelling algorithms are frequently used in CITC research (e.g., Jiang et al. 2021; Miranda et al. 2022b; Syed and Silva 2023). Therefore, topic modelling algorithms offer a timely and relevant opportunity to demonstrate how researchers can achieve methodological rigour when using algorithms in the research process. Due to the increasing interest, a few scholars have also offered topic modelling tutorials to guide novice researchers (Debortoli et al. 2016; Schmiedel et al. 2019). However, none of these explicitly discuss the methodological rigour. Moreover, in their recent literature review of topic modelling, Günther and Joshi (2020) highlight the need for methodological rigour in topic modelling-based studies. Overall, there is little conversation about how to achieve methodological rigour with topic modelling. Therefore, the objective of this paper is to offer guidance on achieving methodological rigour in topic modelling.

By illustrating the application of the structural topic modelling (STM) algorithm (Roberts et al. 2013) to a corpus of blockchain academic article abstracts, I discuss how to establish methodological rigour in each step of the process. STM was selected from a range of topic modelling algorithms because it is well suited for extracting prominent topics from a collection of short text such as article abstracts (Roberts et al. 2014). I propose a set of guidelines to help researchers in ensuring methodological rigour in topic modelling studies. The contributions of this paper are twofold. Firstly, I underscore the need for critical assessments of research practices used in algorithm application to minimise the potential threats and increase the trust in research. Second, through the STM illustration and the set of guidelines, I contribute to topic modelling literature and more broadly to methodological rigour in IS research. The paper offers a more nuanced understanding of the application of topic modelling algorithms in research. I join the emerging dialogue on methodological rigour in CITC research. The guidelines are helpful, especially for novice scholars applying topic modelling and editors and reviewers who engage with topic modelling manuscripts. While these guidelines are intended for topic modelling algorithms, they can also be applied to other machine learning (ML) algorithms, including supervised ML. Their applicability, however, requires critical evaluation based on the research context.

The remainder of this paper is organised as follows. First, I offer an overview of topic modelling algorithms. Second, I describe a three-stage process and illustrate how to achieve methodological rigour in topic modelling. Third, I discuss the guidelines. Finally, I conclude with contributions, limitations, and future work.

# 2 Topic Modelling

In this paper, I primarily focus on the probabilistic topic modelling algorithms. Probabilistic topic modelling refers to a suite of machine learning algorithms that employ statistical methods to discover thematic structures ('topics') from massive collections of documents inductively (Blei 2012). In the simplest form, topic modelling algorithms generate a set of topics shared amongst the entire collection of documents. Each document exhibits the same set of topics but in different proportions (Blei 2012). Each topic is represented through a set of words, where each word has a probability of belonging to a topic. In practice, topics are described using the most probable words and assigning meaningful labels. Topic modelling algorithms provide these useful probabilistic measures with substantial predictive power (e.g. word distributions of topics) for us to further analyse document corpus.

Latent Dirichlet Allocation (LDA) developed by Blei et al. (2003) is notably the widely used topic modelling algorithm (Blei 2012). Variations of LDA and other topic modelling algorithms are developed to extend the basic topic extraction capabilities. For example, the Structural Topic Model (STM) algorithm (Roberts et al. 2013) is a combination and extension of multiple topic models including LDA. STM can incorporate document-specific metadata (e.g. publication date of each document) into the model to provide strong topical inference. It allows us to understand relationships between topics and document-level metadata. Dirichlet-multinomial regression topic model (Mimno and McCallum 2008), dynamic topic model (Blei and Lafferty 2006), and author-topic model (Rosen-Zvi et al. 2004) are designed with similar capabilities for varying purposes. Neural topic models (NTM) that leverage deep neural networks are increasingly gaining popularity. Instead of using the bag-of-words model to represent documents, NTMs use contextualized representations (Abdelrazek et al. 2023). LDA2VEC (Moody 2016) and ETM (Dieng et al. 2020) use word embeddings with LDA. BERTopic allows us to use any pre-trained language model for word embeddings (Grootendorst 2022). These topic modelling algorithms convert documents into word embedding representations before clustering them into topics. Each of these topic modelling algorithms has its strengths and weaknesses. Therefore, the choice of specific algorithms is context-dependent.

Scholars have used topic modelling algorithms to examine a variety of phenomena and develop theories. For example, Miranda et al. (2022b) used topic modelling to unpack the diversity and coherence paradox of innovation discourse. Syed and Silva (2023) identified social movement frames and mobilization structures through topic modelling. Furthermore, several scholars have provided methodological guidance on topic modelling. Debortoli et al. (2016) and Schmiedel et al. (2019) offer step-by-step tutorials using an illustrative example of how to apply topic modelling. They discuss different stages of the process (e.g., data collection, data preparation), challenges, strategies, and best practices. Palese and Piccoli (2020) offer guidance specifically on evaluating the interpretability of topics through human judgment. Although these papers provide helpful guidance, the emphasis on systematically achieving methodological rigour is limited. In a recent literature review of topic modelling in management and IS research, Günther and Joshi (2020) find scholars treat best practices as standardized procedures without making context-specific adjustments. Such practices can potentially undermine the plausibility of knowledge claims derived based on the patterns yielded from topic modelling. Against this backdrop, in this paper, I specifically focus on methodological rigour in topic modelling applications.

# 3 Ensuring Rigour in the Application of Topic Modelling

Although there are different views on research rigour, scholars generally agree that rigour must be ensured in all aspects of the research process including research design, data collection, data analysis, and theory development (Dubé and Paré 2003; Soliman and Siponen 2022). Ensuring rigour includes demonstrating established quality standards, stringent application of research methods, and conducting rigorous analysis and interpretation of data (Soliman and Siponen 2022). Moreover, researchers must evaluate and justify numerous decisions made in each stage of the research process as these decisions can influence the quality of the research output (Venkatesh et al. 2013). In this paper, I focus on the rigorous application of algorithms which is a crucial part of CITC research as this research relies on the patterns surfaced through algorithms to develop theory.

The application of algorithms in a research setting is an iterative process from preparing data to building multiple models and evaluating patterns that surfaced to select a final model. Figure 1 visualises this iterative process in a simplified manner. This three-stage iterative process is a simplification of the schematic steps presented by Shmueli and Koppius (2011) as their schematic covers the complete list of activities a researcher involves in when using algorithms in research context. Although Shmueli and Koppius (2011) present the steps in linear order, in reality, model building is rarely a linear process. Thus, it is helpful for researchers to have an agile mindset as they need to iterate the three-stage process multiple times. This is because the nature of the data, decisions made during the data preparation stage, and choices of input parameters have a greater influence on the model output. Although many tend to think algorithms generate patterns, in practicality, patterns are human constructions. Therefore, it is important to ensure rigour in each stage of the process.

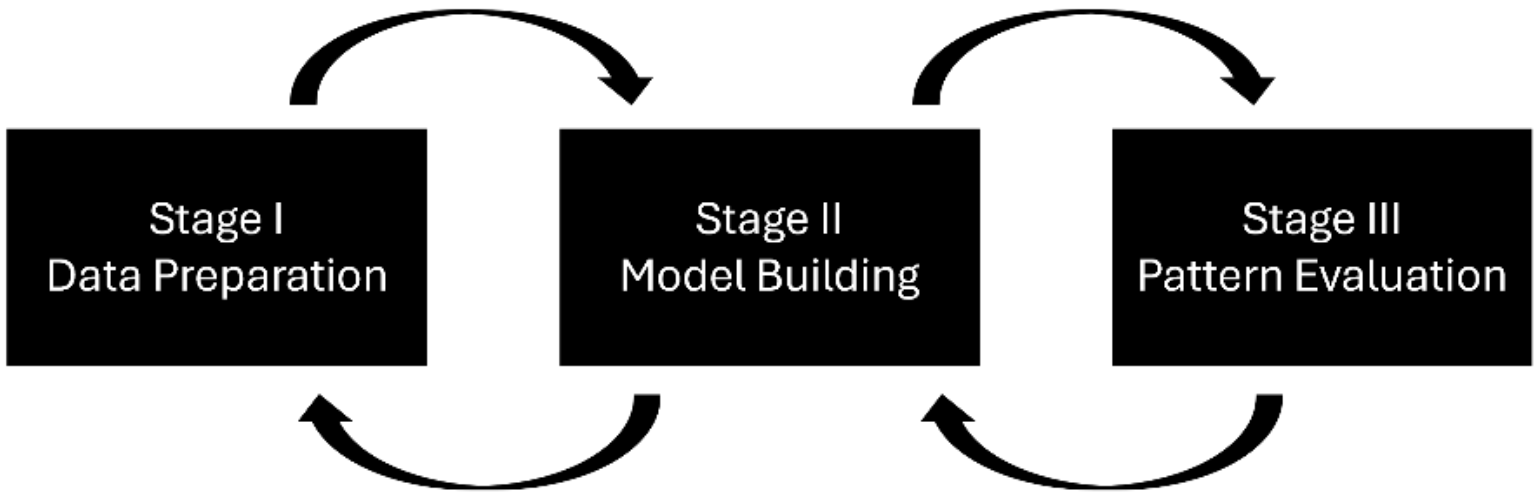

*Figure 1: Three-Stage Iterative Process*

While data selection choices are important for the complete research project, for the scope of this paper, I only focus on three stages: data preparation, model building, and pattern evaluation. These three stages are more important to establish methodological rigour. Using an academic article abstract dataset, I illustrate the most important points where a researcher needs to establish rigour in topic modelling applications. I use the three-stage process depicted in Figure 1 to guide my illustration of the rigorous application of topic modelling. In each stage, I discuss the choices researchers have, how these choices influence the algorithmic output, and justifications for the choices made.

## 3.1 Empirical Setting

### 3.1.1 Dataset

The dataset chosen for illustration is comprised of academic article abstracts about blockchain. I chose blockchain due to its popularity and relevance as a research topic within research communities. However, it should be noted that my purpose is to demonstrate how to apply a topic modelling algorithm rigorously in a research setting. I searched articles from ABI/INFORM Global via ProQuest using the keyword 'blockchain' for the period from 2008 to 2021. ABI/INFORM Global includes scholarly publications covering a wide variety of business-related disciplines including prominent IS journals such as MIS Quarterly. I set the start year to 2008 which is the birth year of Bitcoin because the concept of blockchain originated through bitcoin (Nakamoto 2008). I also filtered the search results to only retain journals and conference articles in the English language. This resulted in 1528 articles, and I then downloaded the abstracts of these articles.

### 3.1.2 Topic Modelling Algorithm

I chose the Structural Topic Modelling (STM) algorithm to illustrate how researchers can achieve methodological rigour. The dataset chosen for the illustration is comprised of short documents which are article abstracts. STM is well suited for extracting prominent topics from a collection of short text as one of its earliest applications by seminal authors is extracting topics from open-ended survey responses (Roberts et al. 2014). The core innovation of STM compared to other probabilistic topic modelling techniques is that it lets the user build additional information into the topic model itself. This additional information can be either document-level metadata such as published date or user-defined data such as document category. As such, it allows us to observe how the prevalence and content of topics vary along any document-level covariate (Roberts et al. 2014). From a technical point of view, STM can be best understood as running a regression on to topic proportion over document-level metadata. STM is not only capable of detecting topics and topic change over user-defined covariates, but also inferring topic correlations and deciding the optimal number of topics which is a common practical challenge when applying topic models. STM offers a software package in R (statistical computing programming language) with a set of useful functionalities including visualization (Roberts et al. 2019).

## 3.2 Stage I: Data Preparation

I used the R programming environment and relevant packages throughout the topic modelling process. First, I removed the duplicate records from the dataset. Then I examined 20 abstracts to obtain an initial understanding of the data and data quality problems (Debortoli et al. 2016). Next, I converted all the text into lowercase. Through the initial exploration, I observed that some article abstracts contain phrases such as 'abstract', 'design/methodology/approach', and 'originality/value'. Therefore, I created a custom word list and removed those phrases that are part of the document structure of an abstract. I also removed the term 'blockchain' as it is our search keyword. Through the initial examination of the abstracts, I also identified a list of collocations of words such as 'information technology', 'smart contracts', 'supply chain', and 'distributed ledger technology'. In the next step, I concatenated these words because the meaning is derived together rather than from the individual words. Then I removed stop words using the 'tm' package (Feinerer 2020). I also added a few custom stop words through our exploration of the abstracts. Next, I removed special characters and numbers. As extra whitespaces may be retained after this step, I removed all the white spaces including the leading and trailing blank spaces. Then I performed lemmatization using the R package 'textstem' to harmonize different forms of the same word. I did not apply stemming as it can distort the meaning in a given context (Evangelopoulos et al. 2012). Finally, using the 'stm' package itself, I performed tokenization and removed words that are shorter than three characters. This resulted in 1520 documents with 4944 terms and 118236 tokens.

## 3.3 Stage II: Model Building

STM requires three decision parameters (also called hyperparameters) that need to be set before running the algorithm[1]: initialization type, maximum number of iterations, and number of topics. I set the initialization to "Spectral" because spectral initialization outperforms the other two available types, LDA and random initialization (Roberts et al. 2019). While the default setting for the maximum number of iterations is 500, I set it to 1000 to provide sufficient time (in terms of iterations) for model convergence. The most crucial decision in topic modelling algorithms in general is setting the number of topics as this parameter affects the quality of the patterns that surfaced (Blei et al. 2003; Roberts et al. 2019). Although model accuracy may be better with a higher number of topics, it may not be useful for human interpretation and, thus should not solely rely on those (Blei 2012; Chang et al. 2009). There is no right answer to the number of topics for a given document collection as this decision also depends on the goal of the research study (Blei 2012; Chang et al. 2009; Roberts et al. 2019). Therefore, a researcher needs to find a balance between the model's predictive accuracy and interpretability that suits the research goal when it comes to evaluating this decision.

To find the optimal number of topics (denoted by K) that is appropriate for our dataset, I used the 'searchK' functionality in STM. This function allows us to compare multiple models for a given range of K and provides several model diagnostic metrics to examine the inference quality such as held-out log-likelihood, residuals, semantic coherence, and exclusivity. Both held-out log-likelihood and residual analysis assess the predictive accuracy of the model by measuring how well the learned patterns hold for unseen data (Taddy 2012; Wallach et al. 2009). The higher the held-out log-likelihood and the lower the residuals imply better model accuracy. However, both held-out log-likelihood and residual analysis do not assess the explanatory goals or the interpretability of the topic model (Chang et al. 2009). Therefore, to assess interpretability, STM provides two metrics namely semantic coherence and exclusivity (Roberts et al. 2019). Semantic coherence is based on the idea that high-probability words tend to co-occur together frequently, thus it assesses whether a topic is internally consistent (Roberts et al. 2019; Roberts et al. 2014). However, achieving high semantic coherence is comparatively easier by having a few topics with very common words (Roberts et al. 2019). In addition, semantic coherence does not discriminate the topics that are alike (Roberts et al. 2014). Exclusivity helps assess how well topics are different from each other (Roberts et al. 2014). Therefore, it is recommended to use both semantic coherence and exclusivity for evaluating the interpretability of models.

First, I performed several experimental rounds of topic models to refine the lists of collocation of words and custom stop words. This is often a necessary step to get a better understanding of the data quality issues and how the algorithm performs (Debortoli et al. 2016). Then I evaluated 12 models to determine the number of topics by assessing the metrics related to model accuracy and interpretability (K=5, 10, 15, 20, 25, 30, 35, 40, 45, 50, 55, 60). Figure 2 presents the model diagnostics generated via STM. The models with 10, 15, and 20 topics have the highest held-out likelihood. Semantic coherence is highest for the model with 5 topics and shows a decreasing trend in general. Residuals are lower from 35 to 60.

[1] Refer to Roberts et al. (2019) for detailed discussion.

Considering the models with high held-out likelihood and semantic coherence I chose 10, 15, and 20 models for the next evaluation. I excluded model 5 as it has the highest residual value. Furthermore, within the range in which residuals are lower, I included 35 and 45 models as they have relatively higher held-out likelihood and semantic coherence. I did not use the lower bound metric as I was only interested in assessing model accuracy through held-out likelihood and residuals, and interpretability of models through semantic coherence in this step. I chose five models 10, 15, 20, 35, and 45 to further assess the algorithmic interpretability through semantic coherence and exclusivity of individual topics.

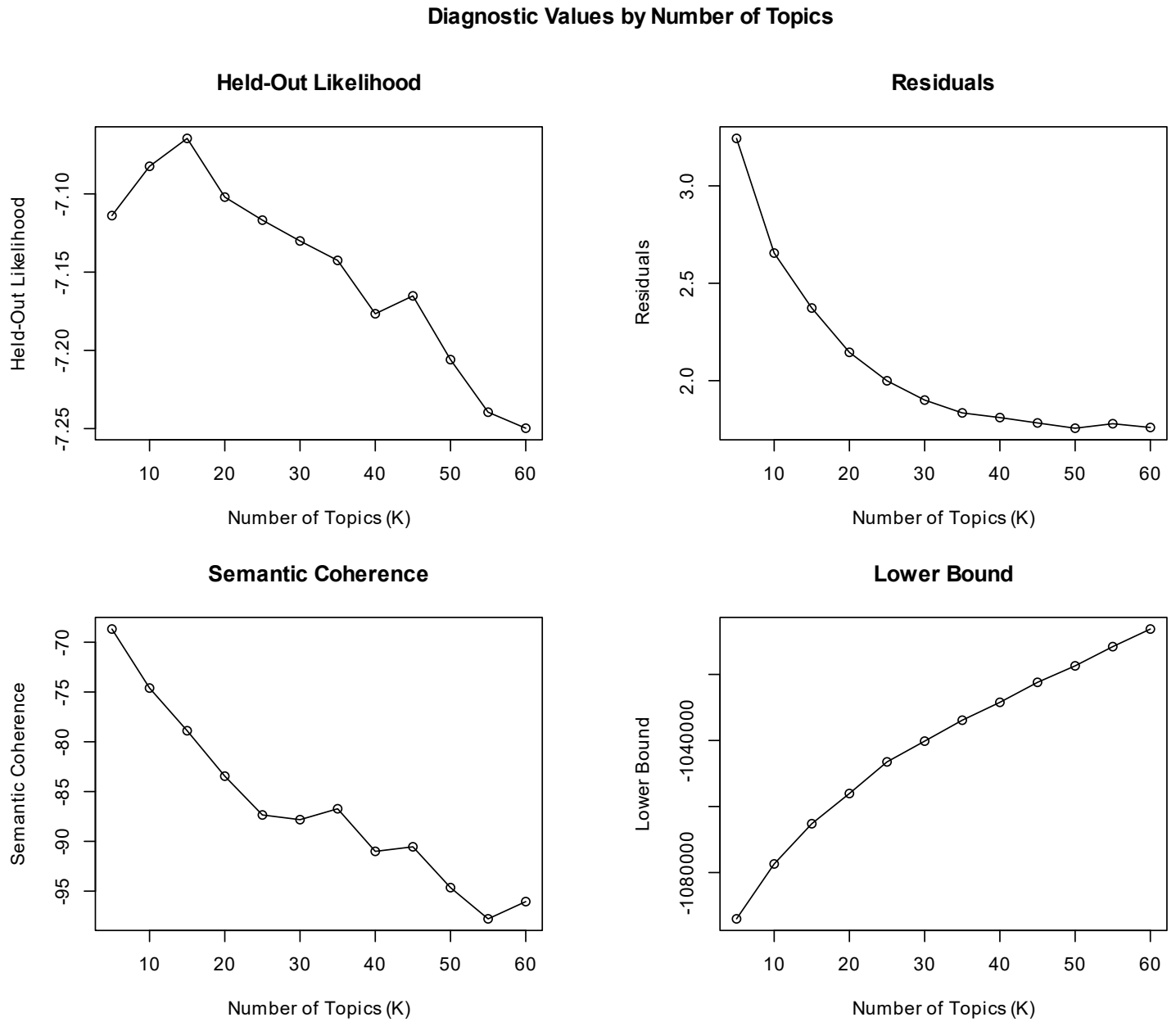

*Figure 2: Diagnostic Metrics (Semantic coherence, exclusivity, residuals)*

Figure 3 presents the semantic coherence and exclusivity of each topic for the five models. The models that have topics with higher semantic coherence and exclusivity have better interpretability. In other words, when most of the topics lie in the top right corner of the graph is an indicator of having better interpretability. Models 10 and 45 have several topics with relatively lower exclusivity and semantic coherence. Therefore, I selected models 15, 20, and 35 for further evaluation.

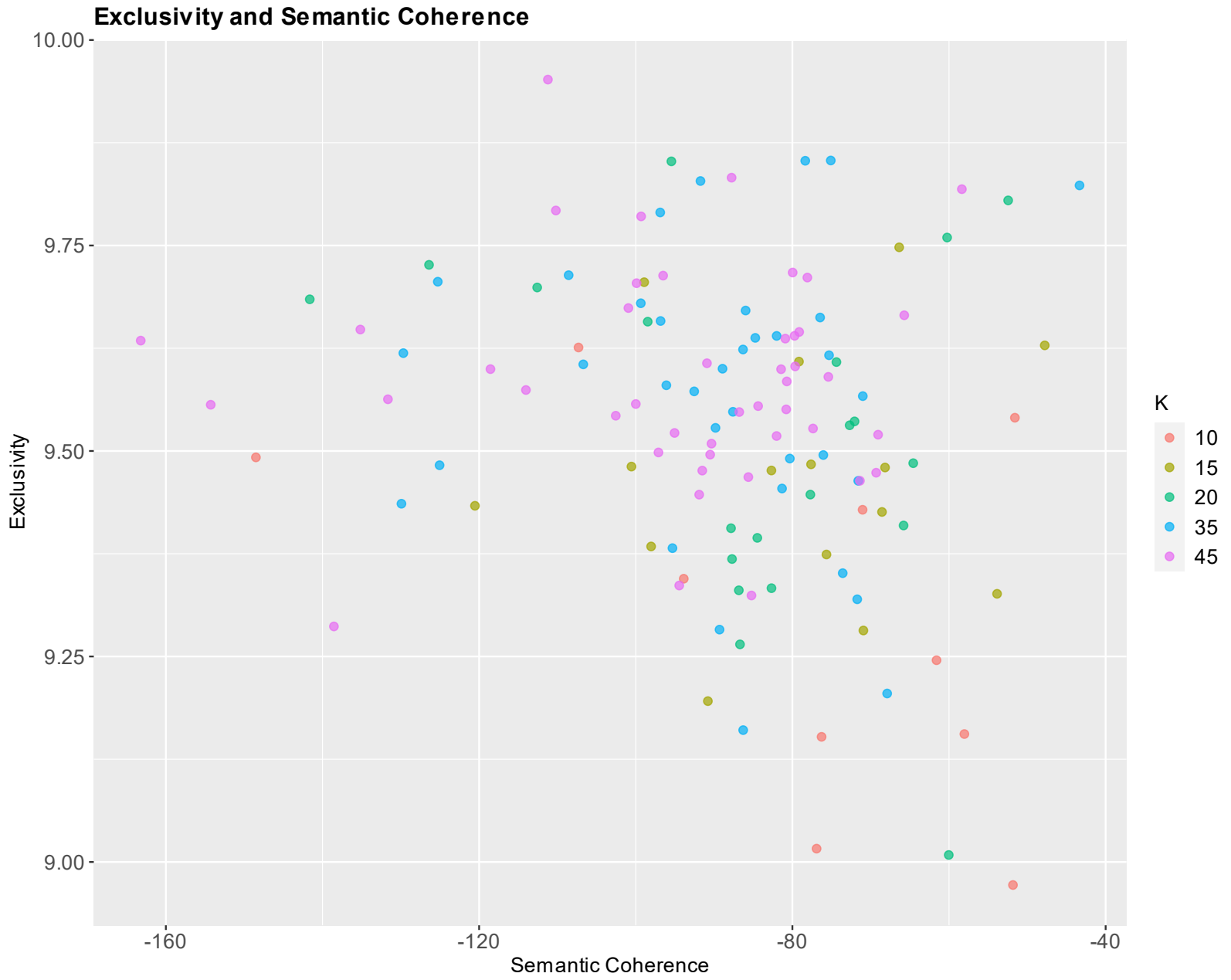

*Figure 3: Semantic Coherence and Exclusivity Scores*

## 3.4 Stage III: Pattern Evaluation

While semantic coherence and exclusivity provide us with an assessment of how coherent and different individual topics are, they do not provide an indicator of how meaningful individual topics are and the respective topic assignment to article abstracts. Therefore, to further assess the interpretability of models, I performed a manual evaluation task. I used the word intrusion and topic intrusion tasks designed by Chang et al. (2009). While the word intrusion task assesses "*whether a topic has human-identifiable semantic coherence*" (Chang et al. 2009, p. 3), the topic intrusion task assesses "*whether the association between a document and a topic makes sense*" (Chang et al. 2009, p. 3). The goal of the human coder is to identify an intruder word inserted into a given topic and an intruder topic of a given article abstract which associates least with the abstract. The results of these two tasks are then evaluated through the two respective metrics: model precision and topic log odds. Refer to Chang et al. (2009) for further details about these tasks.

I used the 'oolong' R package which offers an easy-to-use implementation of the word and topic intrusion tasks. As the dataset is about blockchain, I needed a coder who has sufficient knowledge about blockchain. Therefore, I selected a research assistant who is well-versed in both the business applications and technical aspects of blockchain. Two coders (external coder and I) performed these tasks for the chosen topic models 15, 20, and 35. The word intrusion tasks were performed per topic for all the selected models. For the topic intrusion tasks, I configured 10 cases per model. That is each coder performed 10 topic intrusion tasks per model. Figures 4 and 5 depict snippets of word intrusion and topic intrusion tasks. Table 1 presents the results of word and topic intrusion tasks. The model with 20 topics has the highest model precision and the second-highest topic log-odds scores. Therefore, I chose this model as the final model.

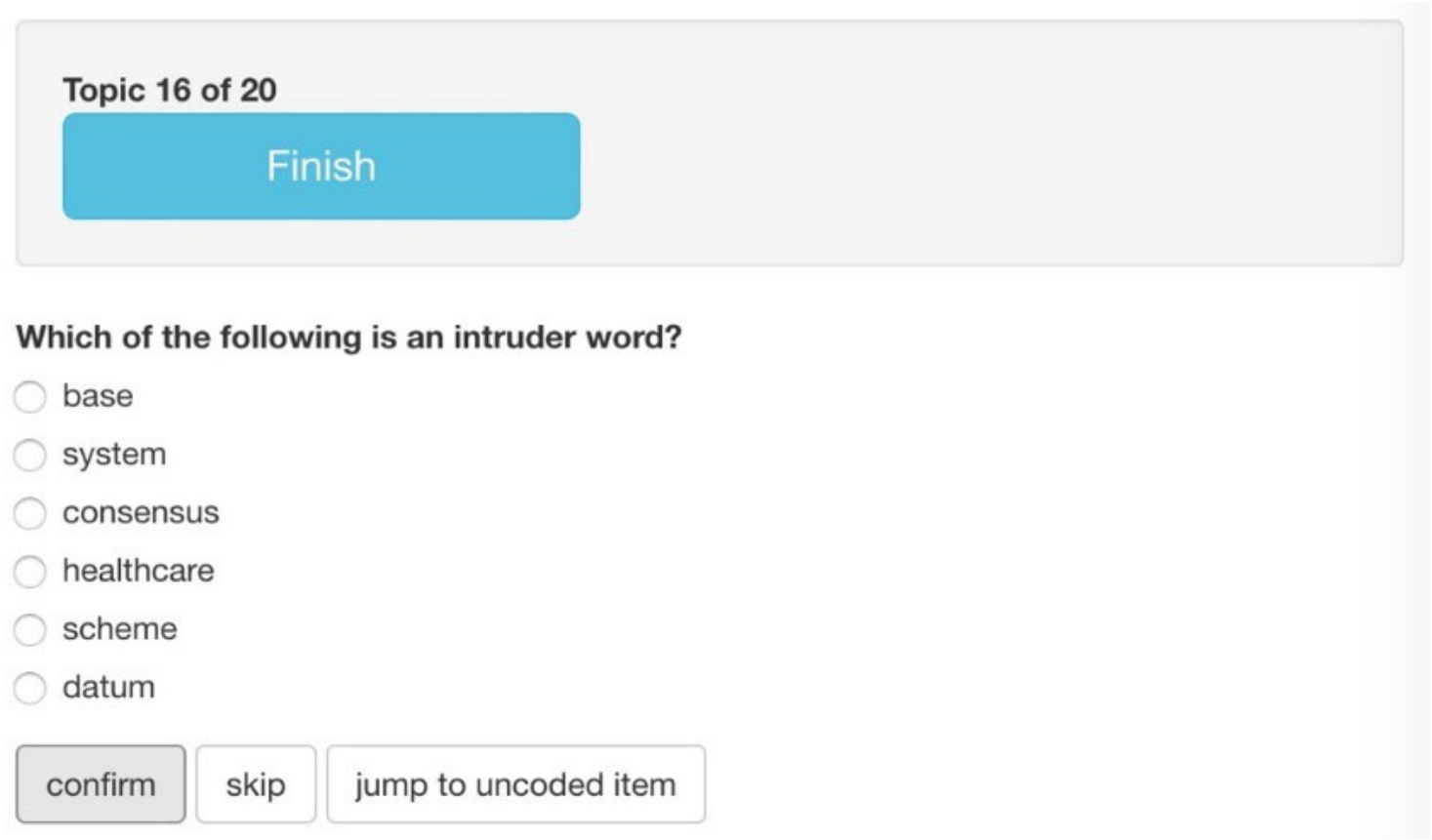

*Figure 4: Word Intrusion Task*

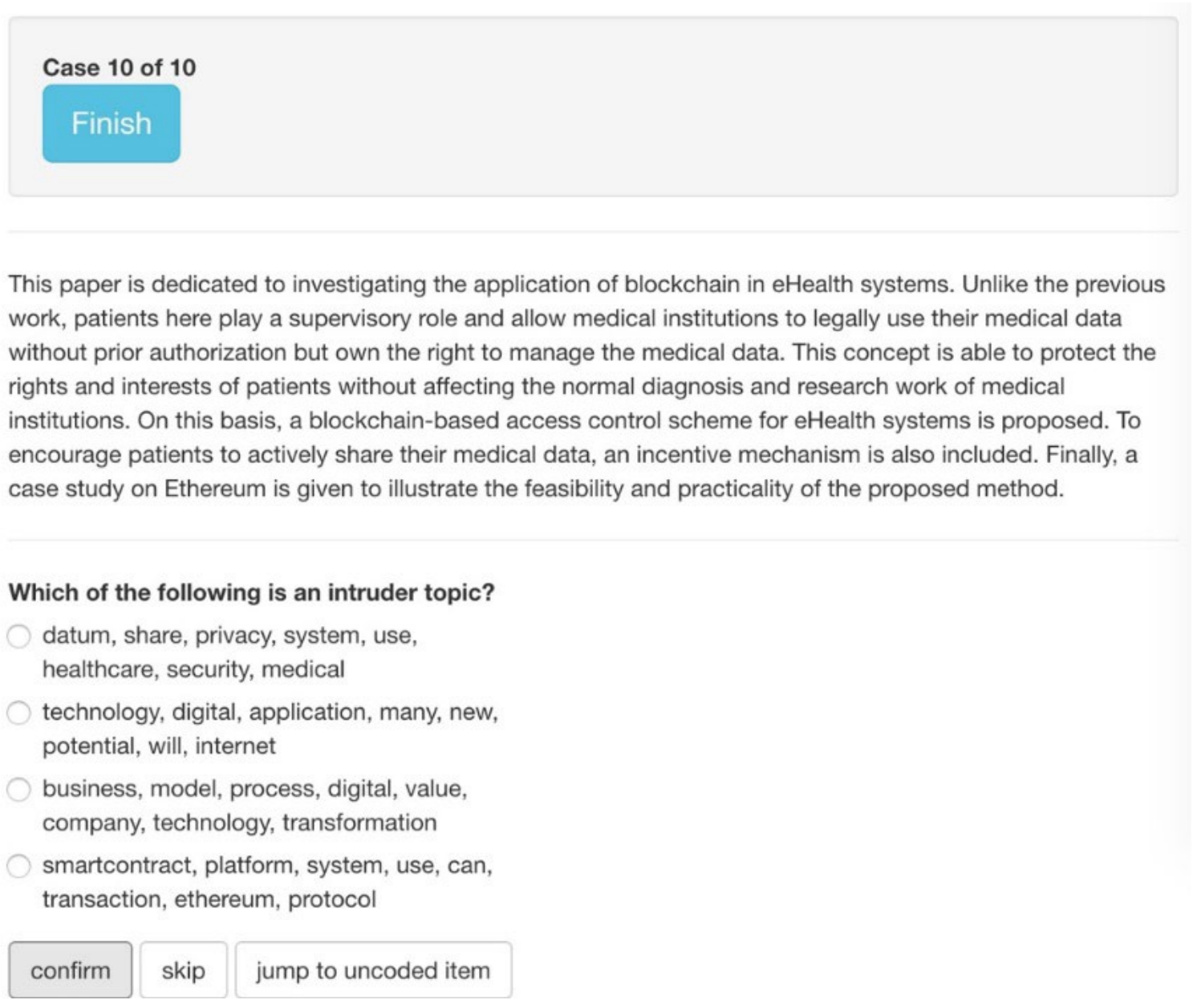

*Figure 5: Topic Intrusion Task*

| Model | Model Precision | Topic Log Odds |
|---|---|---|
| Model 15 | 0.57 | -1.57 |
| Model 20 | 0.68 | -1.09 |
| Model 35 | 0.50 | -1.01 |

*Table 1. Topic and Word Intrusion Metrics*

The next crucial task is to provide human-inferred labels for the topics. As this is a highly subjective task, it is also important to assess the quality of these interpretive labels. Therefore, I designed an evaluation task to assess the quality of human interpretability of the topics. Which topics to be included and excluded in the study is dependent on the goal of the study. For example, a scholar who adopts topic modelling to conduct a literature review or historical analysis may include all the topics. As my purpose here is to demonstrate methodological rigour, I only chose the top 10 topics with the highest probability (refer to the Appendix for all the topics). For the topic labelling task, an external coder with the five most probable words and the ten most probable article abstracts corresponding to each of the top 10 topics. Subsequently, the external coder and I independently assigned labels to each topic. The final labels for the top 10 topics were determined through a joint discussion with the external coder.

| Topic | High Probable Words | Topic Label |
|---|---|---|
| Topic 6 | research, literature, paper, study, review | Literature Review |
| Topic 16 | datum, system, scheme, base, consensus | Consensus Algorithms |
| Topic 17 | iot, datum, security, privacy, smart | IoT Solutions |
| Topic 15 | distribute, system, ledger, transaction, decentralize | Distributed Ledger Solutions |
| Topic 3 | technology, will, artificialintelligence, new, digital | Digital Technology Revolution |
| Topic 18 | cryptocurrency, ico, money, currency, exchange | Cryptocurrency Investments |
| Topic 13 | technology, industry, development, research, study | Blockchain Adoption |
| Topic 10 | business, model, process, value, technology | Business Model & Process Innovation |
| Topic 12 | supplychain, system, use, food, can | Supply Chain Solutions |
| Topic 19 | public, governance, economic, new, social | Socio-political & Socio-economic Impact |

*Table 2. Top 10 Topics with Corresponding Top Five Words and Topic Labels*

# 4 Discussion

In this section, I offer a set of guidelines on how researchers can establish rigour when they apply algorithms. Although these guidelines are based on topic modelling algorithms, they can be applied to other algorithms (e.g., supervised machine learning algorithms) with context-specific adjustments. The first six guidelines correspond to the three-stage process (Figure 1). Additionally, I discuss two guidelines related to the reporting[2] of computational procedures. Table 3 presents these guidelines which I elaborate on next.

## 4.1 Data Preparation

### 4.1.1 Understand Which Aspects of Data Can Influence the Patterns

Topic models can be used to surface topics in a variety of data such as social media posts, academic articles, interview scripts, and email archives. Thus, the required data preparation activities differ based on the nature of the data. For example, social media posts contain hashtags and mention symbols that require elimination during the data preparation process. Whether to eliminate only the '@' symbol or the accounts mentioned in the data is a decision that needs to be made based on the study objective. If there are accounts that were frequently mentioned in the data, then that can even result in a separate topic. If the researcher is new to topic modelling, these decisions are impossible to be made upfront. Thus, it is important to experiment with the data and undertake multiple iterations of the three-stage process (Figure 1). Even for an experienced researcher, if they are new to the type of data, they may require multiple iterations of the process to identify how the data influences the model output and which activities are appropriate for data preparation. Importantly, researchers need to approach the application of algorithms in research settings with an agile mindset, i.e., with a willingness to iterate the research process.

[2] Reporting is part of the complete research process irrespective of the research genre. However, I have not included it in my three-stage process as this paper focuses on rigour in the application of algorithms.

| Stage/Aspect | Guidelines |
|---|---|
| Data Preparation | Understand which aspects of data can influence the patterns |
| | Assess the suitability of existing standard procedures |
| Model Building | Understand how hyperparameters influence model accuracy and interpretability |
| | Balance between model accuracy and interpretability |
| Pattern Evaluation | Explore existing evaluation tasks |
| | Design context-specific evaluation tasks |
| Reporting Computational Procedures | Report for transparency |
| | Report to help build the community |

*Table 3. Guidelines for Achieving Methodological Rigor*

### 4.1.2 Assess the Suitability of Existing Standard Procedures

Günther and Joshi (2020) find that some topic modelling studies have followed prior practices without giving sound justifications. Researchers must critically assess the suitability of standard practices or procedures followed by prior studies against their research context before applying them. For example, the text mining practice of stemming may not be useful for data sourced from social media platform X (former Twitter). Often people shorten the words to maximise the content written in a tweet due to its 280-character limitation. Moreover, people use lots of hashtags that may have a specific meaning in a particular context but are not part of a standard language vocabulary. Therefore, standard pre-processing steps such as stemming, and lemmatization should be used with careful assessment of its suitability. In addition, some existing practices may not be relevant for certain topic modelling algorithms. For example, new topic modelling algorithms such as ETM do not require the standard text mining practice of removing stop words (Abdelrazek et al. 2023). It is important to provide sound justifications for each of the decisions researchers make along the way.

## 4.2 Model Building

### 4.2.1 Understand How Hyperparameters Influence Model Accuracy and Interpretability

One needs to understand hyperparameters (i.e., input parameters like the number of topics) well enough to make decisions about them. Different values specified for model parameters influence the predictive accuracy and interpretability of the model. Also, the data may influence the values specified for hyperparameters. One way to gain a better understanding of how these parameters influence the output is to perform sensitivity analysis by experimenting with different values for model parameters. It helps to understand how the algorithm optimizes predictive accuracy and how the quality of patterns changes accordingly. This understanding will then help to identify which metrics are appropriate to assess model accuracy and interpretability for the chosen topic modelling algorithm. It is important to note that other various metrics are also available, beyond the metrics I've used (e.g., held-out log-likelihood and coherence). However, researchers must not treat these metrics as standards. Rather, they should select based on a clear understanding of the algorithm's sensitivity to model parameters.

#### 4.2.2 Balance Between Model Accuracy and Interpretability

Many algorithms originate from the computer science discipline, the goal of which is often to prioritize algorithmic performance. This is not always compatible with the goal of the broader social science and management disciplines, which is often to formulate interpretive explanations about a phenomenon (Hofman et al. 2021). The model with the highest predictive accuracy may not always be the most useful. Therefore, striking a balance between achieving methodological rigour in model accuracy and model interpretability is mostly up to the researcher and depends on the goal of the study. For example, if the goal is to develop theoretical models using prediction, then a researcher may focus more on rigour in algorithmic performance whereas if the goal is to develop comprehensive theoretical explanations, then a researcher may focus more on rigour in algorithmic interpretability (Debortoli et al. 2016; Shmueli and Koppius 2011; Shrestha et al. 2021). Therefore, the balance between model accuracy and interpretability should be guided by the research goal.

### 4.3 Pattern Evaluation

#### 4.3.1 Explore Existing Evaluation Tasks

Researchers can borrow existing evaluation tasks to assess whether the patterns are meaningful and relevant and whether human interpretations of the patterns are plausible. However, it is important to assess the suitability of these tasks and report the justifications. For example, word and topic intrusion tasks are existing evaluation tasks that are specially designed for evaluating the plausibility of topics (Chang et al. 2009). Palese and Piccoli (2020) designed evaluation task to assess topics' interpretability based on human judgment. They created three human-labelled gold-standard sets for hotel reviews and made them available to the research community for reuse. There have been attempts to design topic modelling algorithms such as partially labelled topic models that also generate topic labels (Ramage et al. 2011). Therefore, exploring the available evaluation tasks that are up-to-date, especially in the computer science and social science community will be beneficial. As the text mining community is constantly working in this area, we can expect new developments such as fully labelled interpretable topic models in the future. However, the task of evaluating the meaningfulness, relevance, and plausibility of patterns will continue to persist.

#### 4.3.2 Design Context-Specific Evaluation Tasks

If existing evaluation tasks are inappropriate for a given research context, one possibility is to design evaluation tasks catered to the context. Researchers must ensure the patterns are meaningful and relevant to answer their research question. Thus, this approach provides flexibility to design tasks that can not only assess the human interpretability of the patterns but also assess their meaningfulness and relevance to the research context. For example, Syed and Silva (2023) designed a manual evaluation task focusing on three criteria: semantic validity (whether the model meaningfully discriminates between similar words), lower number of human interpretable topics, and mutual exclusive topics. Another possibility is to use generative AI capabilities to evaluate patterns. For example, one can design an evaluation task to prompt ChatGPT to generate summaries of the topics and evaluate these labels using human experts (Rijcken et al. 2023). There are abundant opportunities for researchers to apply their creativity in designing context-specific evaluation tasks.

### 4.4 Reporting Computational Procedures

#### 4.4.1 Report for Transparency

Being explicitly transparent about the research process is necessary to demonstrate the research rigour because it helps readers (including reviewers) understand why and how the research was conducted. Research transparency increases trustworthiness in research, facilitates replication, and promotes innovation and generativity (Burton-Jones et al. 2021). In general, research transparency refers to the practice of explicitly reporting how the research study is undertaken, specifically the design and execution-related methodological details. Transparency is becoming ever more important for assessing the rigour in algorithmic applications because computational methods are often black-boxed and require specialized skills (Leone et al. 2021). Therefore, researchers should demonstrate methodological rigour through explicit reporting of the computational procedures. Disclosure of computational procedures helps research communities in two ways. Firstly, reporting the systematic procedure applied enables assessing the plausibility of the patterns and research outcomes specifically during the review process. Secondly, a well-documented procedure provides ample guidance to ensure rigour for novice scholars who wish to adopt the same algorithm in future research. This helps future scholars to replicate the procedure similarly and bring further innovation (Leone et al. 2021; Pare et al. 2016).

### 4.4.2 Report to Help Build the Community

There are emerging initiatives within scholarly communities (e.g., Management Science) to provide programming codes and other details of the computational procedures during the publication process (Leone et al. 2021). There exists a plethora of algorithms to date and new algorithms are emerging at an exponential rate. The need to develop a paradigm for training scholars is at the forefront of leveraging these methodological innovations (Lazer et al. 2009). The encoding of know-how knowledge can increase the repertoire of skills needed for scholars who do not have a strong computer science background or programming skills. Therefore, sharing procedures and programming codes helps others to reuse the procedure with less effort and creates a supportive community to accelerate the adoption of algorithms while ensuring their rigorous application. Some scholars have already initiated reporting detailed information about their process of applying algorithms. For example, Syed and Silva (2023) discuss in great detail how they applied the LDA algorithm in their study of social movement.

# 5 Conclusion

Advanced computational algorithms have changed the way we conduct research, with scholars recognizing their potential for theory development. Computationally intensive theory construction (CITC) is an emerging genre of research that leverages algorithms to develop theory (Berente et al. 2019; Miranda et al. 2022a). Central to this approach are computationally derived patterns that underpin the theorizing process. While algorithms offer significant potential, they also present methodological challenges in research. Their inherent opacity complicates scholars' understanding of pattern generation and their suitability for theory construction. The increasing use of algorithms in research, with a lack of transparency and methodological rigour, can undermine trust in research (Günther and Joshi 2020). Yet, the conversation about achieving methodological rigour in the application of algorithms in CITC research is still emerging. Topic modelling algorithms are frequently used in CITC research (e.g., Syed and Silva 2023). In this paper, I attempt to offer guidance on achieving methodological rigour specifically in the application of topic modelling algorithms. By illustrating the application of the structural topic modelling (STM) algorithm (Roberts et al. 2013) to a corpus of blockchain academic article abstracts, I discuss how to establish methodological rigour in each step of the process. The proposed guidelines are useful for novice scholars applying topic modelling and reviewers of topic modelling manuscripts. They can be applied to other ML algorithms, subject to critical assessment based on the research context.

I was motivated to write this paper due to two reasons. First, there is a lack of transparency and methodological rigour in studies that use algorithms as highlighted by Günther and Joshi (2020) as well as my observations as an author and a reviewer. Second, there is a lack of guidance in establishing methodological rigour in CITC research where the algorithms are key because it is still in a nascent stage. Therefore, through this paper, I intend to make two key contributions. First, I highlight the importance of methodological rigour when applying algorithms in research settings, particularly it is crucial when computationally derived patterns are used to develop theory. Second, by illustrating the application of the STM algorithm and offering a set of guidelines, I contribute to topic modelling literature and more broadly to methodological rigour in IS research. Specifically, I contribute to the discourse on methodological guidance in topic modelling (e.g., Debortoli et al. 2016; Hannigan et al. 2019; Palese and Piccoli 2020; Schmiedel et al. 2019). By focusing on the rigorous application of algorithms, I also join the emerging dialogue on methodological rigour in CITC research (e.g., Günther and Joshi 2020; Miranda et al. 2022a). The guidelines are helpful, especially for novice scholars applying topic modelling and for the editors and reviewers engaging with topic modelling manuscripts.

This work is not without limitations. This paper only focuses on the three stages: data preparation, model building, and pattern evaluation. It will be beneficial to develop guidelines for the complete research process such as assessing the appropriateness of the chosen algorithm for answering research questions in a specific research context. For example, if a scholar is interested in applying topic modelling algorithms on question-and-answer (Q&A) content, topic modelling approaches designed for Q&A content (e.g., Zhang et al. 2024) can be more suitable than using LDA or STM. Moreover, data preparation decisions such as handling missing data and standardization can be important for the performance of the model (Choudhury et al. 2021; Shmueli and Koppius 2011). Therefore, future studies can extend this work to include the complete research process to derive a comprehensive list of guidelines. Furthermore, future work can develop and extend guidelines for other kinds of algorithms such as supervised ML, deep learning, and large language models. Finally, with the rapid emergence of new algorithms, developing design principles for a family of algorithms (e.g., supervised ML) can assist research communities in leveraging these algorithms for theory-building endeavours.

# Appendix

Figure A1 presents the complete list of topics that surfaced from the blockchain academic article abstracts through the application of STM.

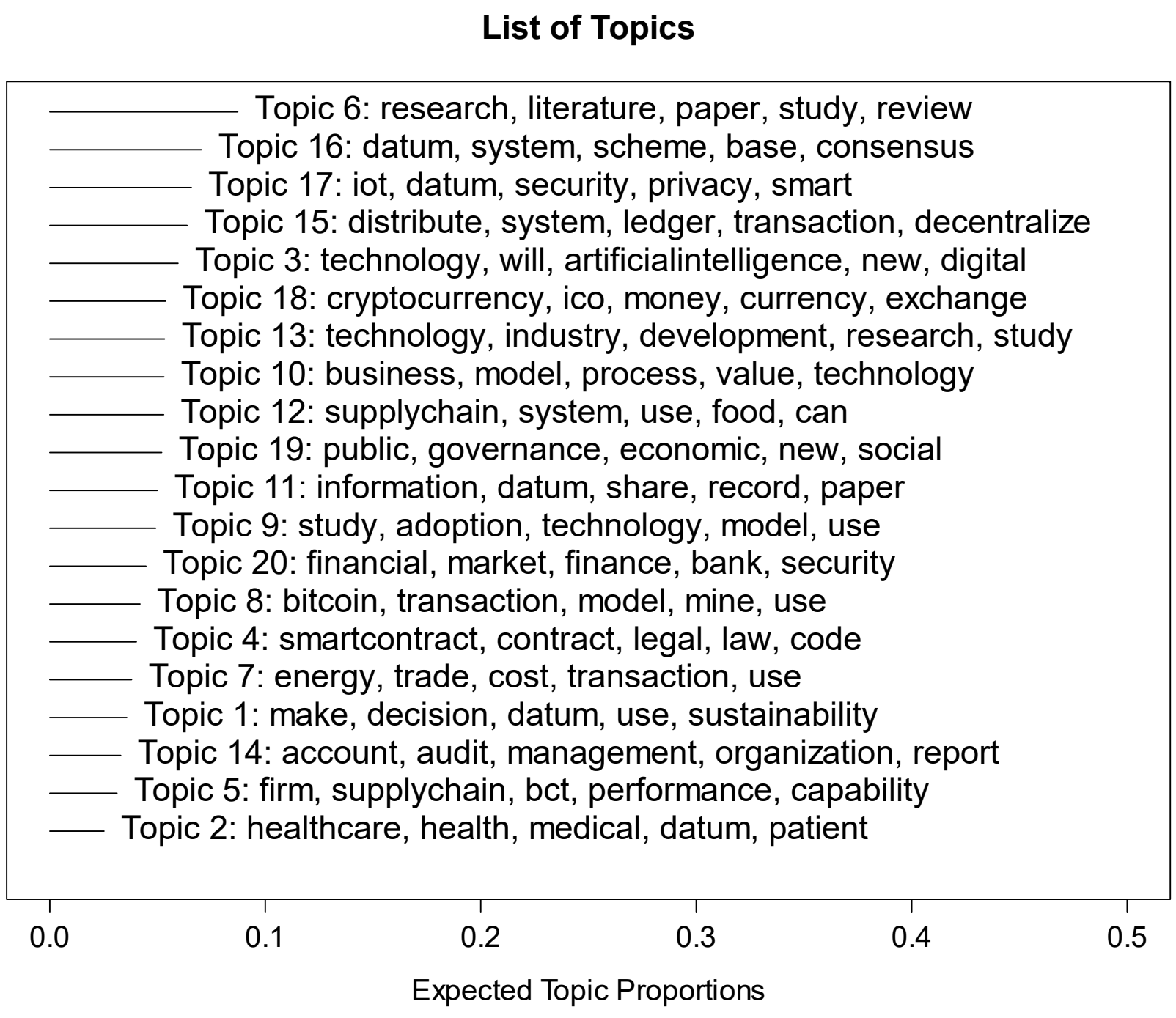

*Figure A1: List of Topics and Mean Topic Proportions*

# Copyright